\documentclass[preprint,12pt]{elsarticle}
\usepackage{amsmath}
\usepackage{enumitem}
\usepackage{comment}
\usepackage{amssymb}
\usepackage{graphicx}
\usepackage{hyperref} 
\usepackage{tabularx}
\usepackage{subcaption}
\usepackage{float}
\journal{Safety Science}

\begin{document}

\begin{frontmatter}



\title{Ergonomic Optimization in Worker-Robot Bimanual Object Handover: Circumventing the Discrete Nature of REBA Scores Using Reinforcement Learning in Virtual Reality}


\author[inst1]{Mani Amani}

\affiliation[inst1]{organization={Computational Science Research Center, San Diego State University},%Department and Organization
            addressline={5500 Campanile Drive}, 
           city={San Diego},
            postcode={92182}, 
            state={CA},
            country={USA}}
\affiliation[inst2]{organization={Department of Civil, Construction, and Environmental Engineering San Diego State University},%Department and Organization
            addressline={5500 Campanile Drive}, 
           city={San Diego},
            postcode={92182}, 
            state={CA},
            country={USA}}
            \author[inst1,inst2]{Reza Akhavian\corref{cor1}}
\ead{rakhavian@sdsu.edu}
\cortext[cor1]{Corresponding author}

\begin{abstract}
Robots can serve as safety catalysts on construction job sites by taking over hazardous and repetitive tasks, while alleviating the risks associated with existing manual workflows. Research on the safety of physical human-robot interaction (pHRI) is traditionally focused on addressing the risks associated with potential collisions. However, it is equally important to ensure that the workflows involving a collaborative robot are inherently safe, even though they may not result in an accident. For example, pHRI may require the human counterpart to use non-ergonomic body postures to conform to the robot hardware and physical configurations. Frequent and long-term exposure to such situations may result in chronic health issues. Safety and ergonomics assessment measures can be understood by robots if they are presented in algorithmic fashions so optimization for body postures is attainable. While frameworks such as Rapid Entire Body Assessment (REBA) have been an industry standard for many decades, they lack a rigorous mathematical structure which poses challenges in using them immediately for pHRI safety optimization purposes. Furthermore, learnable approaches have limited robustness outside of their training data, reducing generalizability. In this paper, we propose a novel framework that approaches optimization through Reinforcement Learning, ensuring precise, online ergonomic scores as compared to approximations, while being able to generalize and tune the regiment to any human and any task. To ensure practicality, the training is done in virtual reality utilizing Inverse Kinematics to simulate human movement mechanics. Experimental findings are compared to ergonomically naive object handover heuristics and indicate promising results where the developed framework can find the optimal object handover coordinates in pHRI contexts for manual material handling exemplary situations.
\end{abstract}



\end{frontmatter}


\section{Introduction}
\label{sec:sample1}
Production and assembly work in industries such as construction still
relies heavily on human workers, causing a decrease in productivity and
safety, in part due to physical stress and ergonomic tension \cite{KRUGER2009628}. Robotics development and research in construction has been a heavily evolving field in recent years, focusing on improving productivity, quality, and safety. However, physical human-robot interaction (pHRI)in construction, particularly with robots having a manipulator robotic arm has received very little attention in such contexts {\cite{RODRIGUES2023104845}}. This is while other industries such as manufacturing have made significant strides in safe pHRI recently {\cite{JAHANMAHIN2022102404}}.
Methods such as the Rapid Entire Body Assessment (REBA), Rapid Upper Limb Assessment (RULA) \cite{mcatamney1993rula}, and Ovako Working Posture Analysis (OWAS) \cite{KARHU1977199} \cite{hignett2000reba} are industry standards for measuring and assessing the ergonomic safety of tasks to prevent Work-related Musculoskeletal Disorderss (WMSDs)  in fieldwork contexts. Many pHRI research also uses these industrial frameworks as a means to account for human safety in collaborative tasks \cite{hamad2023concise}. %add more ref. 
REBA and RULA work based on a scoring system using predefined body angles and postures to determine the ergonomic risk to the participant given the activity. The range for these scores typically spans discrete positive integers. RULA ranges from 1-7 and REBA ranges from 1-15. This scoring regiment creates an easy, fast, and transparent method of manually gauging the ergonomic safety of a task. However, this discreteness makes the scores non-differentiable, making mathematical optimization a difficult task for automated approaches. This creates an issue when trying to autonomously optimize for optimal REBA scores in pHRI tasks. These frameworks are made for easy and transparent intervention as opposed to optimization. To overcome this obstacle, previous works have developed a wide array of techniques. 
Some researchers tackle this issue by collecting data from a pHRI task and then augmenting the dataset using methods such as regression, which closely smoothens the distribution while still maintaining the raw score reading. \cite{yazdani2022dula} This differentiable method can then be used to optimize ergonomic safety in tasks such as object co-manipulation in HRI by defining and optimizing the dynamics of the task. Other studies aim to create predictive techniques such as quadratic approximations \cite{numerical}. Non-gradient-based optimization algorithms such as genetic algorithms have also proved to help improve ergonomic scores in HRI object handover tasks \cite{omidi2023improving}. However, most current research has been focused on small and light objects through one-handed handover tasks. Many of these works also focus on isolated aspects of object handover, such as release time or correction during co-manipulation. These approaches mainly require human data collection and the use of previous task-specific data to model and optimize, while not being focused on practical applicability for the fieldworkers in a cluttered environment such as those seen in construction sites.
%We focus on the global location of object handover in a work environment utilizing Building Information Models (BIM) which can help with tracking work and evaluating the task to help guide the robot through a dynamic job site.
%Current research has shown that human kinematics can be modeled with high fidelity in simulated environments \cite{li2023niki}.
\par
Another challenge in optimization for ergonomics in pHRI is the dependency of the approach on specific physical tasks. Traditional analytical and approximate solutions must account for the varying dynamics of the object and task, requiring frequent reevaluation and adjustments as tasks change. To overcome this, model-free approaches such as Reinforcement Learning (RL) are attractive options. RL operates by assessing the current state to determine an action, with the action's quality judged by a reward function. The learner iteratively selects actions based on a predefined action-selection policy, guided by the reward system. While this approach offers task-independent optimization through statistical methods, its need for unique calibration for each task and participant poses practical challenges for real-world application.
\par
We propose the implementation of an RL algorithm designed to pinpoint the optimal location for bimanual object handover tasks. The existing challenges in this area stem from two main issues: the non-differentiability of the REBA scores, which hampers the convergence towards the optimal solution, and the impracticality of gathering data from robots and humans due to constant object position perturbations during training. To address these issues and evaluate the effectiveness of our RL-based methodology, we have developed a high-fidelity 3D virtual environment. This setup simulates the dynamics of pHRI in human-robot object handover scenarios with great detail. %<ATYBE. 
Rapidly advancing in safety training and evaluation and robotics research, Virtual Reality (VR) tools offer innovative ways to study and optimize pHRI without the constraints of the physical world \cite{wang2018critical}. Current Inverse Kinematics (IK) developments have created high-fidelity pose and shape estimation of a parametric humanoid model \cite{li2021hybrik}. Unity is a versatile game engine that supports robotics simulation and integration through the ROS\# plugin \cite{UnityROS2023}. Additionally, packages like Unity ML-Agents \cite{juliani2020} provide a robust framework for conducting robotics simulations within Unity, which could be extended to pHRI testing and workforce training. There have been several previous works that use Unity as a tool for pHRI ergonomic optimization \cite{unity1} and worker ergonomic evaluation in VR settings has already been a widely-researched topic in the field of construction \cite{akanmu2020cyber}.
\par
With an RL approach, we create a task-agnostic collaborative robot training routine that as long as human kinematics are representative of real-life actions, the model can find optimal locations for personalized object handover. Another advantage of this methodology is that the need for differentiable scores has been eliminated; the RL algorithm will get the live REBA scores from the worker's digital twin inside the simulation and optimize for the ideal configuration. Once the learner has converged to the optimal position, the framework is ready to apply this in real-time with a simple transformation of dimensions and coordinates. The main contributions of this paper are as follows:
\begin{enumerate}
    \item Designing a novel task-independent method of finding the best ergonomic location for object placement in bi-manual pHRI object handover in an industrial setting utilizing a fast-converging RL routine.
    \item Presenting a VR framework to train a model for any given task tailored to the physical attributes of the worker collaborator before the model's deployment.
    \item Introducing alternative approaches and circumventions towards ergonomic optimization derived from currently established frameworks for further validity.
\end{enumerate}

\section{Methodology}
\subsection{Q-Learning}
Q-Learning is a model-free RL algorithm for any finite Markov Decision Process (MDP) \cite{ModelFreeMArkov}. The independence of Q-Learning from mathematical models of the environment or the specific problem at hand makes it particularly suitable for HRI object handover ergonomic optimization. This eliminates the need for modeling dynamics or approximations that can have the potential for inaccuracies. The foundation of the model is based on the Bellman Equation \cite{barron1989bellman}, as expressed by Equation (\ref{eq:Bellman}). Where \(s\) is the state, \(a\) is the action, \(R(s,a)\) is the reward, $\gamma$ is the discount factor, \(P(s' \mid s,a)\) represents the transition probabilities and \(V(s')\) is the value of the next state.
\begin{equation}
    V(s) = \max_a \left( R(s, a) + \gamma \sum_{s'} P(s' | s, a) V(s') \right)
    \label{eq:Bellman}
\end{equation}
The Q-Learning equation is as follows:
\begin{equation}
 Q_{\text{new}}(s,a) = (1 - \alpha)Q_{\text{Current}}(s, a) + \alpha \left( r + \gamma \max_{a'} Q(s', a') \right)
    %CHECK WITH CODE
    \label{eq:QLEARN}
\end{equation}
where \(\alpha\) is the learning rate and \(Q(s,a)\) represents the Q-value of the state.
The Q-value illustrates the quality of the action that is taken given the current state. The action is defined by a previously defined action policy such as epsilon-greedy \cite{wunder2010classes} or softmax \cite{song2019revisiting} action policies.
\par
 The softmax action policy chooses an action using Boltzmann distribution and explores all possible actions in every visited state. The alternative candidate actions to be chosen are ranked according to their value estimation.  It chooses action \(a\)  on the \(t\)th  iteration with the probability given by: 
\begin{equation}
    \xi(t) = \frac{e^{\frac{Q_i(t)}{\tau}}}{\sum_k e^{\frac{Q_k(t)}{\tau}}}
    \label{eq:boltzman}
\end{equation}
where \(\tau\) is the temperature \cite{syafiie2004softmax}. As $ \tau \rightarrow 0$, the policy chooses exploitation over exploration, since the actions with lower values have a higher opportunity to be selected, and a higher \(\tau\) tends to pick actions with a higher value. We chose to keep a high \(\tau\) since the reward mechanism is based on a discrete score and it is harder to differentiate action qualities to a significant extent. We decrease the temperature as the score decreases past the score threshold of 5 in a step-wise fashion by 0.1 units, and increase \(\tau\) by 0.1 in the case that the score worsens. In the case of this study, the learner's action set has 6 DOF, moving in a 3D space:
\[
f_{\text{action}}(x, y, z, \text{action}) = 
\begin{cases} 
(x + 1, y, z) & \text{if action } = 0 \\
(x - 1, y, z) & \text{if action } = 1 \\
(x, y + 1, z) & \text{if action } = 2 \\
(x, y - 1, z) & \text{if action } = 3 \\
(x, y, z + 1) & \text{if action } = 4 \\
(x, y, z - 1) & \text{if action } = 5
\end{cases}
\]

\subsection{VR and Inverse Kinematics}
Calculating REBA scores and ergonomic assessment inside a VR environment has been the subject of previous studies \cite{dias2021use}.
Rigging is a procedure used in skeletal animation for representing a 3D character model using a series of interconnected digital bones \cite{lugilde2022biarc}. This technique enables us to reproduce human kinematics at a high level. This framework will take the body measurements such as limb proportions, height, and width to generate a humanoid 3D model. While there is software that generates 3D models automatically, the model can be created manually using modeling software such as 3D Max and Blender. Once the model is created, we can rig the model to ensure accurate kinematics and mechanics to represent the worker in a virtual environment. 
\par
\begin{figure}[h!]
    \centering
    \includegraphics[width=0.7\linewidth]{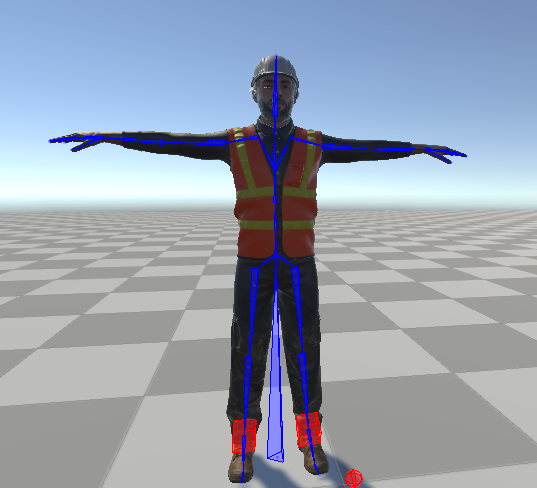}
    \caption{Rigging of the 3D Worker Humanoid Model }
    \label{fig:rigging}
\end{figure}

Previous works have also used Unity as a robotics engine as a means to develop robotic control and perception \cite{RobotEngine}. Our framework also uses the same philosophy in using Unity as a tool for both robotic control and visual representation of pHRI collaboration for increased transparency and multi-disciplinary accessibility. The most recent updates and developments have made this technique viable for seamlessly integrating robotics and game engines \cite{coronado2023integrating}. For our purposes, we utilize the Navigation heuristics, IK solvers, and Animation Rigging features provided by Unity (see Figure \ref{fig:rigging}).
\par
IK is one of the main techniques for granular control over robotic effector manipulation \cite{goldenberg1985complete}. Even though animating human biomechanics through IK is a difficult task \cite{aristidou2018inverse}, there have been promising results in determining pose and ergonomic evaluation in pHRI using 3D simulators and IK \cite{rapetti2023control}. For a generalized postural ergonomic analysis, an accurate humanoid skeletal IK framework will suffice, and other dynamics such as muscular mechanics can be ignored.
\subsection{REBA}
Rapid Ergonomic Body Assessment or REBA is a standard industry-wide metric for overall posture assessment for industrial tasks. REBA is a reliable quantifiable method to evaluate work-related MSDs in the construction industry  \cite{yu2019joint}. The framework defines a from 1-15 for the scores. However, this distribution does not apply to every task. The nature of the task and how a human engages with it will affect the score distribution \cite{rabbani2020ergonomic}. Because REBA is a step-wise linear function, it is non-differentiable. Previous works have overcome this obstacle by introducing methods such as regression neural networks \cite{yazdani2022dula}, as a sum of weighted polynomials \cite{poly} or a continuous linear function \cite{xie2023improving}. This augmentation by the nature of approximation will introduce uncertainty and errors to the process and they mostly have not been tested to confirm their scientific validity in applied settings. 
\par
REBA scores are calculated to joint and body angle ranges which are associated with a score depending on the ergonomic risk. The current state of the art mainly uses image processing as a means of calculating scores for HRI and ergonomic intervention purposes. Other methods such as using inertial measurements \cite{zelck2021combining} and motion capture technologies \cite{yunus2021implementation} have also gained traction. In this framework, the joint angles are directly calculated through the digital bones embedded into the 3D model using the established calculation regiment. This feature gives us exact measurements of joint angles based on the REBA framework as an extension. The proposed framework at this point does not utilize live REBA reading since the relatively optimal position for the specific task is acquired in advance. Once the optimal ergonomic position relative to the worker is acquired, the robot is equipped with a robust policy to ergonomically handover the object at the desired location provided the global position is known. However, during a real-world implementation, it would be prudent to calculate real-time ergonomic metrics and fail-safes to ensure worker safety. 
\par

 \begin{figure}[h!]
    \centering
    \includegraphics[width=0.7\linewidth]{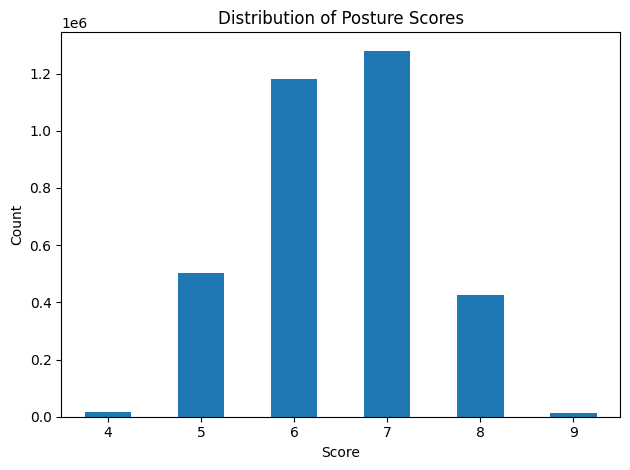}
    \caption{Distribution of Postural Score given a bi-manual HRI Object Handover in the assigned boundary}
    \label{fig:dist}
\end{figure}.

\section{Optimization}
\begin{figure}
    \centering
    \includegraphics[width = 1\linewidth]{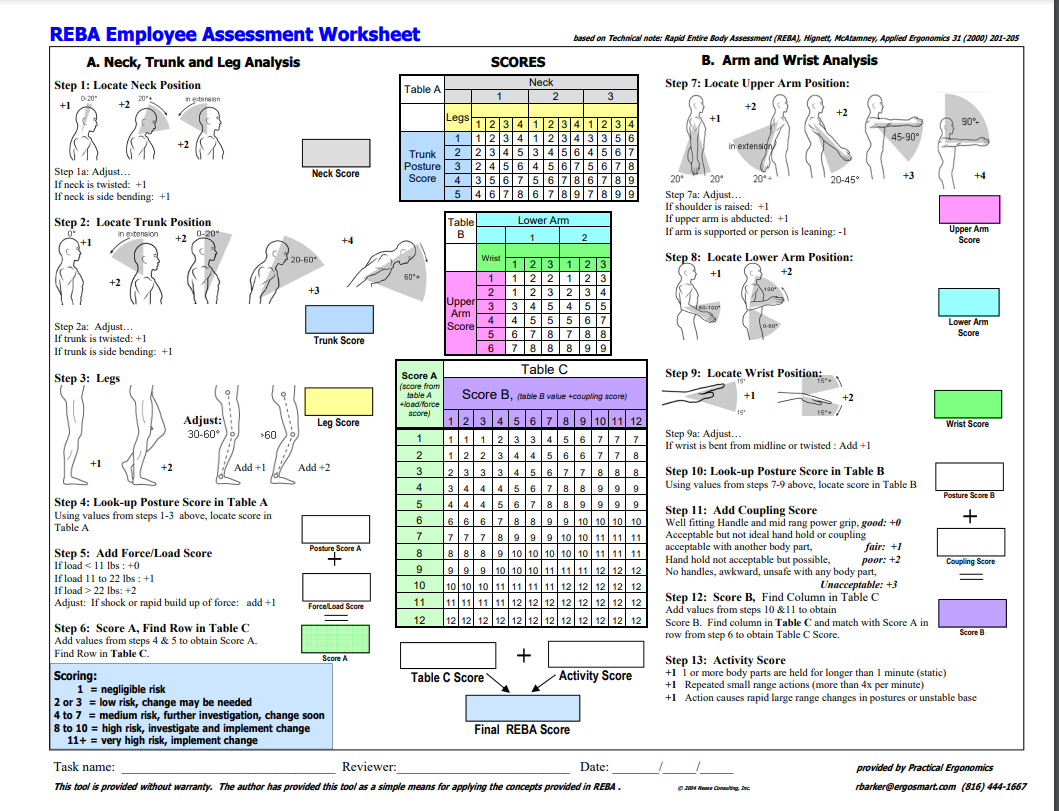}
    \caption{REBA Assessment worksheet}
    \label{fig:worksheet}
\end{figure}

We chose to optimize for an auxiliary score based on the actual REBA calculations which we will call \textit{postural scores} over the final REBA scores in the virtual simulation. The calculation of REBA consists of collecting limb and joint angles which then are processed through a predefined criterion to calculate each score. These scores are then tallied up and filtered utilizing two separate calculation tables into two different categories:
\begin{enumerate}[label=\Alph*.]
    \item Analysis of the neck, trunk, and legs
    \item Analysis of the arms and wrists
\end{enumerate}
Using these two \textit{postural scores}, we can calculate the final REBA score using a third table. However, these tables introduce an additional simplification and discreteness to the distribution. As it can be seen in Fig \ref{fig:worksheet}, the final REBA score can be unchanged even when one of the two scores changes. For example, when the score for category A is 8 and the score for category B is 6, any increase in the value of score B between 6 and 10 would result in the same final REBA score. However, the ergonomic tax has increased even though the actual limb scores for that body category have worsened. This is expected since REBA was not built for precise and accurate ergonomic analysis, especially in VR. However, REBA was created to give an easy, fast, and transparent tool for industry practitioners to measure potential ergonomic harm in industrial settings\cite{REBAbad}. The transformation tables also add discreteness to the distribution. Since our protocol utilizes a reward mechanism based on the improvement of an ergonomic score, the less sensitive score will prove as a harder goal to optimize for. With the postural scores, smaller changes in the positions would result in better or worse scores, helping guide the RL scheme to converge to the optimal position faster. We hypothesize that increased granularity would increase our chances of achieving the true optimal position more reliably. We also argue that the postural score is representative of the true REBA score. The postural score is based on the same score definitions given by REBA for its calculations, meaning it does not perform any different calculation that would be in contradiction with the REBA framework. Furthermore, the relationship between these scores and the REBA score is monotonic, meaning REBA will not decrease as the postural score increases. We further demonstrate that our scheme does improve ergonomic positions by calculating true REBA scores in our experiments. 

The Unity-based simulation incorporates a 3D humanoid model of a construction worker, which is sourced from Mixamo—a platform offered by Adobe that facilitates the creation and rigging of animations and 3D models. This model's body proportions and sizes are adjustable and can be easily rigged with the help of Blender to accurately replicate individual human movements. Once the 3D model is fully rigged, it is imported into Unity. Within Unity, we have designed an IK framework for the digital twin, specifically designed for bi-manual object handovers. This framework leverages Unity’s native functions and is further refined with custom C\# code written by our team to enhance the IK system's precision and adaptability. The lower body and leg movement is controlled by proprietary code and Unity's IKFootSolver. All components and clashes are then handled by a rig controller script. For all other body shapes and sizes, the IK target and hint positions and rig weights will need to be aligned and adjusted according to the dimensions of the model. A postural score calculator script was created using the joint angles of the model following the guidelines defined by the framework. Since the only destination a robot can control and project towards is the position of its end effectors, each end effector would either need to be averaged to the theoretical center of mass of the object or solve and optimize for each effector separately. We chose to optimize for the center of the end effectors using the equation (\ref{eq:cm}). 
\par 

To find the optimal ergonomic score, we utilize a Tabular Q-learning methodology given by equation (\ref{eq:QLEARN}) with a softmax action selection given by equation (\ref{eq:boltzman}). A solution space for the RL scheme was chosen given by the body frame of the 3D model. The size and shape of this distribution set will affect the distribution of the postural scores and by extension the convergence quality and speed. We chose a boundary given by arm extension in the \(z\)-axis, from knee height to the top of the head for the \(y\)-axis and shoulder width for the \(x\)-axis. The reward mechanism utilizes an inverse quadratic relationship of the ergonomic score given by equation \ref{eq:reward}. By opting for an action step size of 0.003 units, this solution space gives us about 3.4 million solutions for the Q-learner to explore. Per Figure \ref{fig:dist}, we can see that most of the distribution belongs to Posture scores 6-7, and only 0.004\% of the solution space belongs to the score 4, which coupled with the discrete nature of the distribution, renders optimization and RL objectives extremely difficult. Recording every possible score for every position took approximately $\sim$3 hours to calculate. REBA, which the calculated postural score is a subset of, and most other rapid ergonomic scores do not account for human subjective preferences. This creates a possibility of obtaining the same minimum optimal score at multiple different positions, causing the algorithm to converge and stop exploring different locations at sub-optimal locations at the global scale. We decided to incorporate a reward value \ref{eq:cm} for the symmetry of the box about the body contact frame given that humans prefer symmetry \cite{evans2012}. This score is then incorporated back into the initial reward function given by \ref{eq:reward2}
Given the discrete distribution of ergonomic scores, the Q-Learning algorithm is prone to convergence on local minima. This often occurs because, in instances where a new position does not yield a change in the ergonomic score, the algorithm's reward becomes dependent solely on symmetry changes within the unchanged score region. As a result, Q-Learning may settle on suboptimal solutions. To counteract this tendency and promote exploration within the model, we opted to keep the softmax action selection temperature relatively high.
\begin{equation}
    r = \frac{1}{{E^2}}
    \label{eq:reward}
\end{equation}
\begin{equation}
    S = X_{\text{LE}} + (X_{\text{RE}} - X_{\text{LE}})
    \label{eq:cm}
\end{equation}
\begin{equation}
    r = \frac{1}{{E^2}} + S
    \label{eq:reward2}
\end{equation}
\par
After the model identifies the lowest postural score, the relative position or positions are saved into a CSV file ready to be used. In a real-world translation, a simple dimension conversion can be introduced to be able to deploy this framework to pHRI settings. Each unit used in the Unity game engine translates to 1 meter in the real world. This standardized approach allows us to be able to convert the optimal locations that are in the "Unity unit" to real-world dimensions depending on the deployment and sensor readings from the deployed robot.

\begin{comment}
    
\begin{equation}
\begin{bmatrix}
x_{\text{optimal}} \\
y_{\text{optimal}} \\
z_{\text{optimal}}
\end{bmatrix}
=
\begin{bmatrix}
x' \\
y' \\
z'
\end{bmatrix}
-
\begin{bmatrix}
\cos \theta & -\sin \theta & 0 \\
\sin \theta & \cos \theta  & 0 \\
0          & 0           & 1
\end{bmatrix}
\begin{bmatrix}
x - x_{\text{rel}} \\
y - y_{\text{rel}} \\
z - z_{\text{rel}}
\end{bmatrix}
\label{eq:coord}
\end{equation}
\end{comment}
\section{Training Results}
\begin{figure}[H]
    \centering
    \includegraphics[width=0.7\linewidth]{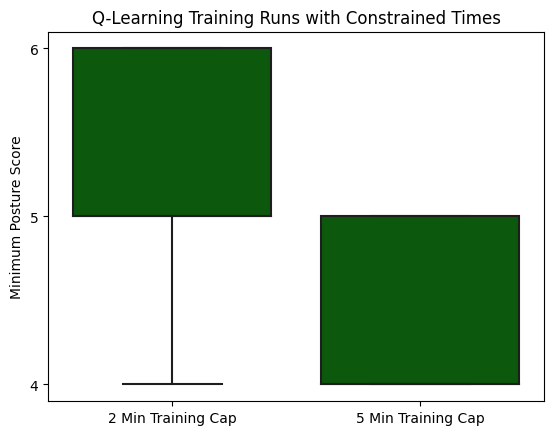}
    \caption{Minimum Postural Score from 10 Separate Training Runs with the Q-Learning Algorithm}
    \label{fig:Training BoxPlot}
\end{figure}

\begin{figure}[h!]
    \centering
    \includegraphics[width=1\linewidth]{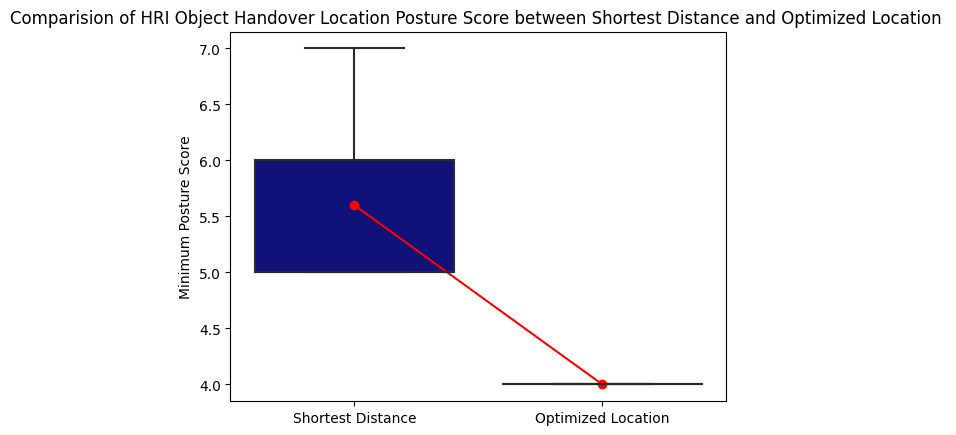}
    \caption{Comparison between the Shortest Distance in the Boundary vs Optimal Location Heuristic}
    \label{fig:Comparission Box Plot}
\end{figure}

\begin{figure}[H]
    \centering
    \begin{subfigure}[b]{0.3\textwidth}
        \includegraphics[width=\textwidth]{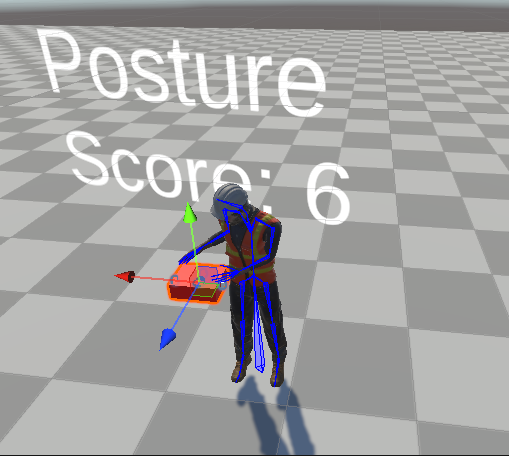}
        \caption{Test Case 2, Shortest Distance }
        \label{fig:image1}
    \end{subfigure}%
    ~ % This adds space between the subfigures
    \begin{subfigure}[b]{0.3\textwidth}
        \includegraphics[width=\textwidth]{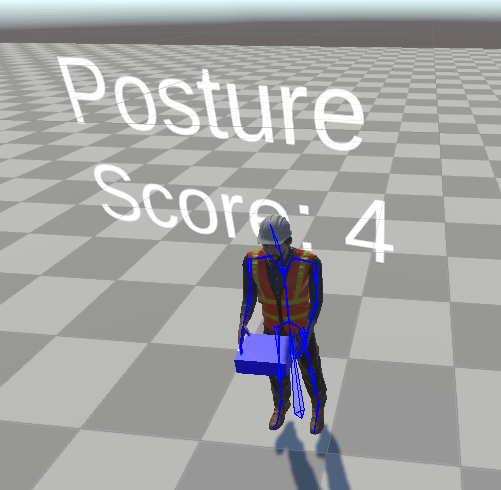}
        \caption{All Test Cases, Optimized Location}
        \label{fig:image2}
    \end{subfigure}%
    ~ % This adds space between the subfigures
    \begin{subfigure}[b]{0.3\textwidth}
        \includegraphics[width=\textwidth]{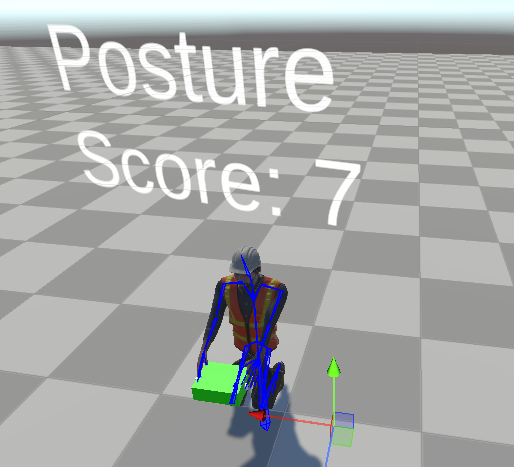}
        \caption{Test Case 4, Shortest Distance}
        \label{fig:image3}
    \end{subfigure}
    \caption{Posture scores based on heuristic and starting location}
    \label{fig:main}
\end{figure}

Due to problems with converging to a local minima, we run the RL algorithm with a time cap. We tested the algorithm with both a 2-minute and a 5-minute cap for 10 iterations. Both time caps were able to achieve the lowest global postural score for the bi-manual object handover location. Results can be seen in Figure \ref{fig:Training BoxPlot}. After acquiring the data, we choose the minimum postural score and the relative position associated with the score. From here we create a script that moves the box to the optimal location. To contrast this with a baseline, another script was created to move the object to the closest distance in the boundary so that the model can reasonably receive the box. This hand-off location has been previously used in the literature, albeit in the context of motion planning and handover approaches \cite{kappler2023optimizing}. We chose 5 locations with different X, Y and Z coordinates outside the model's reach. We then collected the postural score associated with each location. The results show that the optimized approach was able to achieve the minimum postural score at every initial starting coordinate with no variation. This is to be expected since the destination of the framework is independent of the starting coordinates. The shortest distance approach achieved worse REBA at every starting point. The maximum REBA score was 7 and the minimum REBA score was 5. All results are depicted in Table \ref{tab:comparison} and Figure \ref{fig:Comparission Box Plot} and \ref{fig:Training BoxPlot}. Figure \ref{fig:main} shows a sample of the results in the virtual environment. The Unity engine was running on a computer with AMD Ryzen 7 5800 8 Core Processor, 80GB of installed RAM, and an NVIDIA GeForce RTX 12GB 3060 GPU.
\begin{table}[H]
    \centering
    \begin{tabular}{|c|c|c|}
        \hline
        Relative Position & Optimized Location & Shortest Distance \\
        \hline
        (x = 0.028, y = 1.122, z = 1.354) & 4 & 5 \\
        (x = 0.263, y = 1.122, z = 1.354) & 4 & 6 \\
        (x = -0.423, y = 1.122, z = 1.354) & 4 & 5 \\
        (x = 0.028, y = 0.472, z = 0.6) & 4 & 7 \\
        (x = 0.028, y = 2.292, z = 1.354) & 4 & 5 \\
        \hline
    \end{tabular}
    \caption{Comparison of Optimized and Ergonomically Naive methods.}
    \label{tab:comparison}
\end{table}
\section{Experiments}
\begin{figure}[H]
    \centering
    \includegraphics[width =\linewidth]{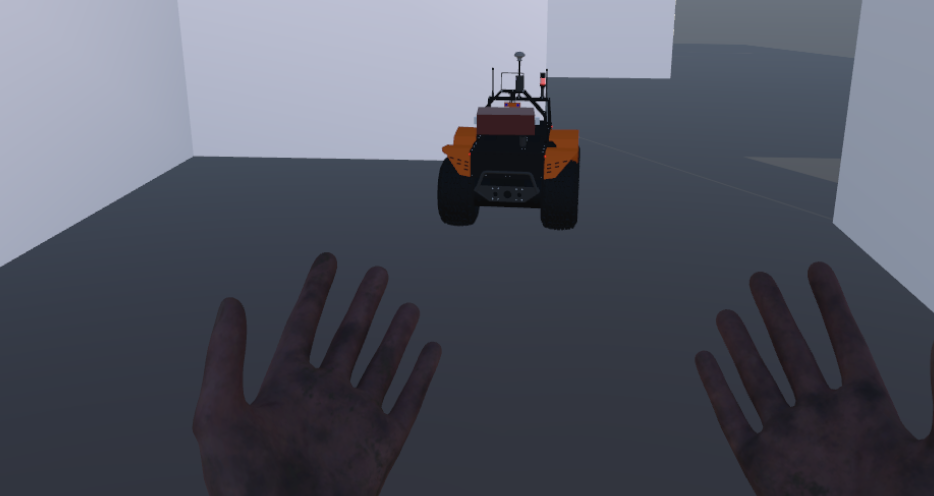}
    \caption{Snapshot of the VR Environment}
    \label{VR}
\end{figure}

\begin{figure}[H]
    \centering
    \includegraphics[width = 0.47\linewidth]{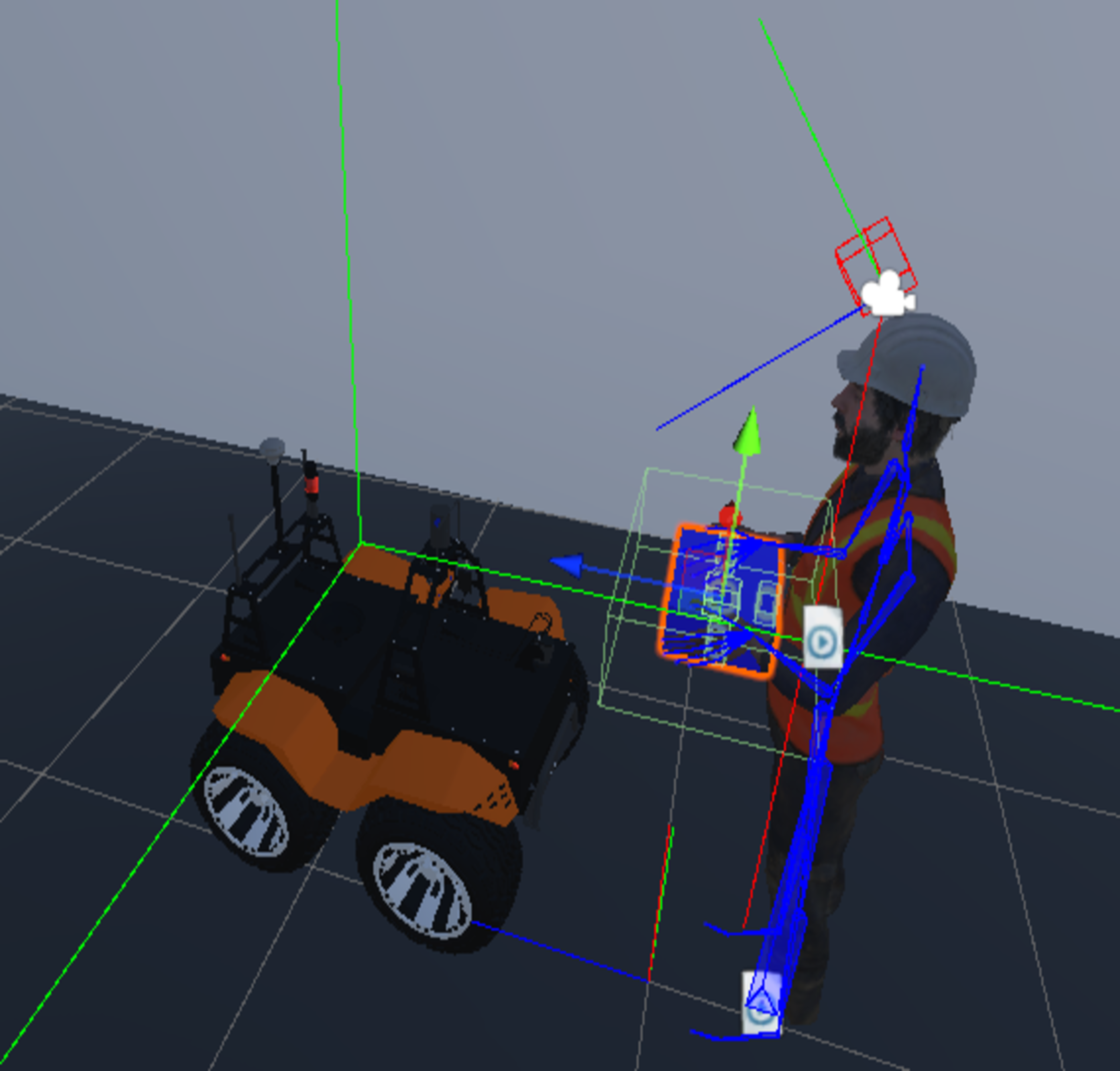}
    \includegraphics[width = 0.47\linewidth]{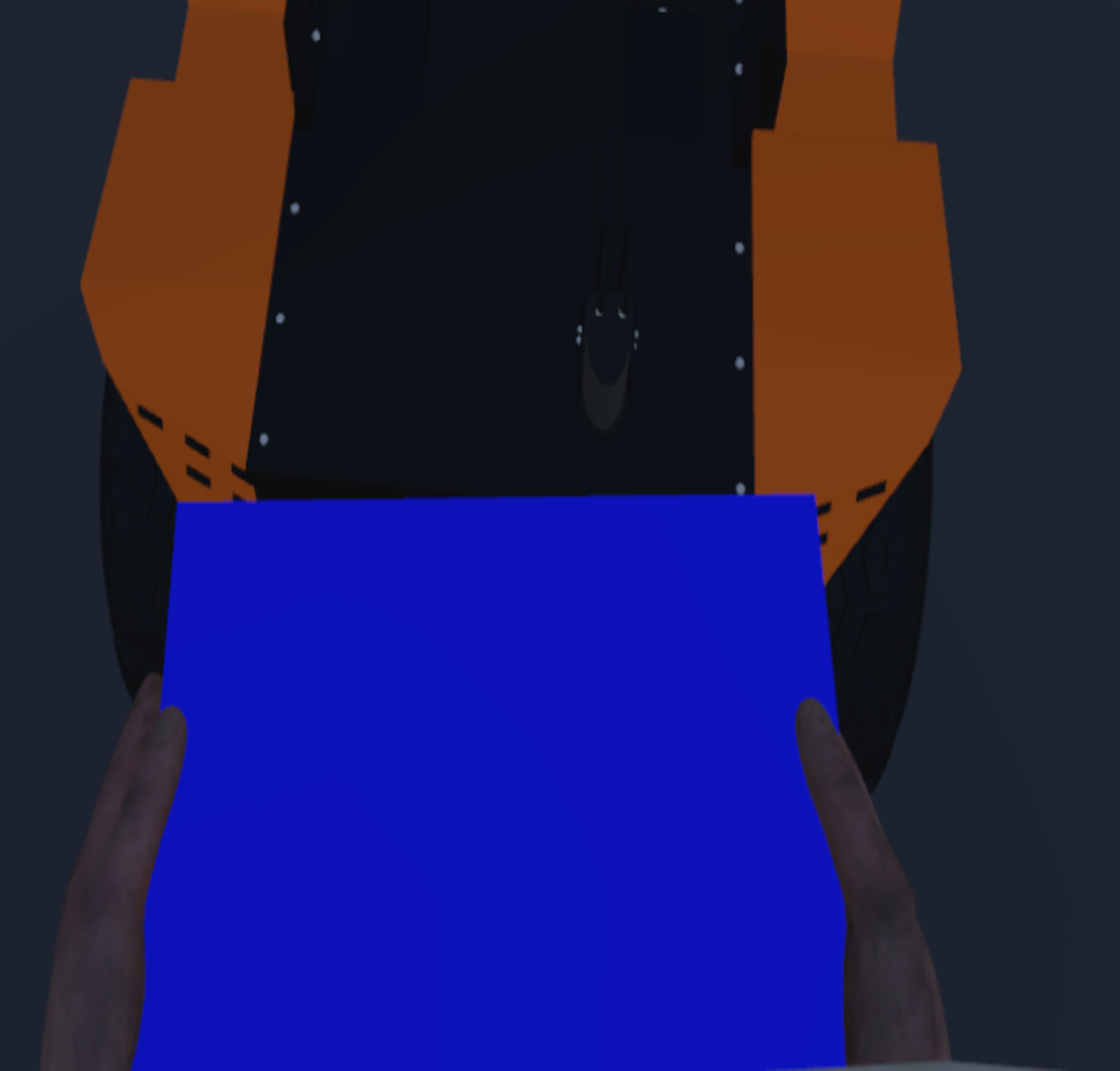}
    \includegraphics[width = 0.47\linewidth]{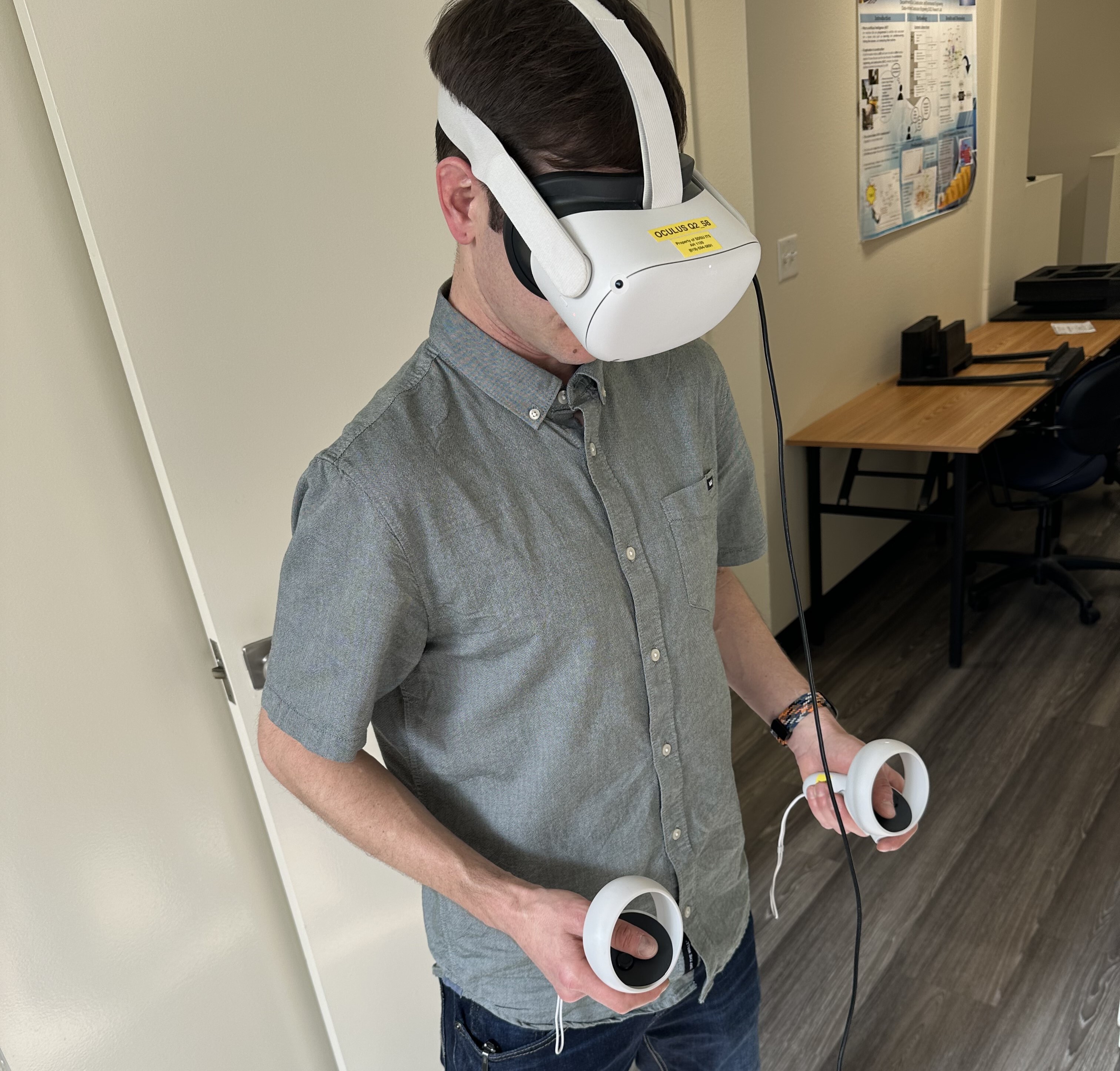}
    \caption{Participant Attempting to Receive the Box in the VR Simulation and the Real-World Body Posture Manifestation}
    \label{evidence}
\end{figure}

To further investigate the efficacy of this optimization regiment, we conducted intra-lab experiments (N=5). A VR scenario was created to test the improvement attributed to the framework. In the scenario, the participants are existing in a room with a box positioned in front of them. We simulate a stationary object handover scenario. A snapshot of the VR environment is depicted in Figure \ref{VR}.
This scenario requires the human to be stationary and the robot to transport the object towards the human for the handover. For this purpose, the participants are instructed to not raise their legs off the ground. However, they were free to move and rotate their body as long as they did not physically move around. Two different object start points were tested. 
\begin{enumerate}
    \item Where the box was directly aligned to the center of the participant's frame
    \item While the box was offset by 1 in-game unit to the left on the x-axis of the participant
\end{enumerate}
After signaling their readiness, the box in the VR simulation is transported toward the participant using the two different (optimized and unoptimized) end locations. After reaching its destination, the box changes color to signal to the participant that is ready for the handoff. The participants are instructed to reach out to simulate grabbing and receiving the box from existing handles on the sides of the object. The participants were asked to hold the position while we took photos of their real-world posture to calculate the manifested REBA score. This procedure is done for both different end locations and start locations. An example of this process is shown in Figure \ref{evidence}

\begin{figure}[H]
    \centering
    \includegraphics[width = 0.45\linewidth]{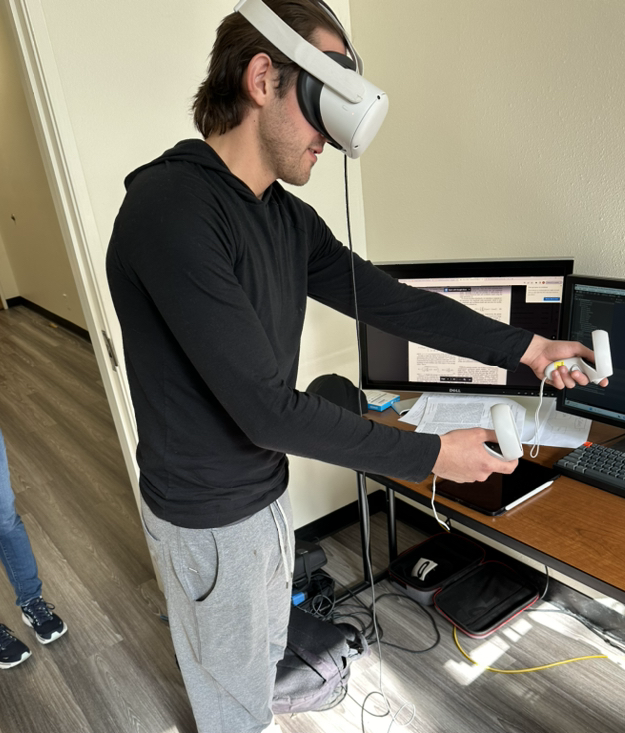}
    \includegraphics[width = 0.45\linewidth]{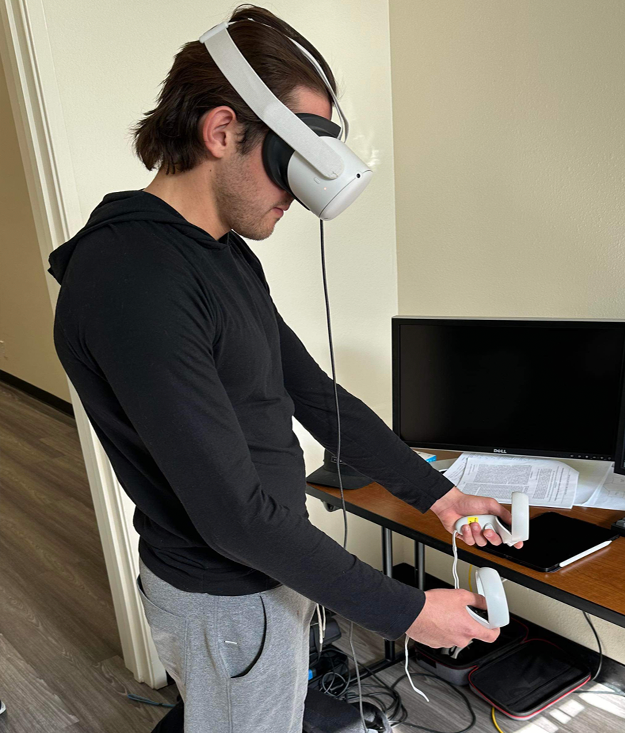}
    \caption{Manifested Postures of the Optimized Framework(left) and Unoptimized Framework (right)}
    \label{fig:blake}
\end{figure}

One shortcoming of assessing ergonomic optimization in VR is the tunnel-visioned worldview that is a byproduct of the headset. The field of view of individuals using VR is significantly less as compared to a natural field of view. This situation is a byproduct of the design and function of the headset \cite{TomsHardware2023}. The reduced field of view can force the users to rotate and change neck position considerably more as opposed to a real-world scenario, which can negatively affect REBA scores due to the higher injury risk associated with neck angles. For consistency and simplicity, we assume the lowest risk neck score between both scenarios to remove the potential confounding factor from our analysis. Figure \ref{fig:blake} shows an example of a recorded data point of the experiment. 
\begin{figure}[H]
    \centering
    \includegraphics[width = 0.675\linewidth]{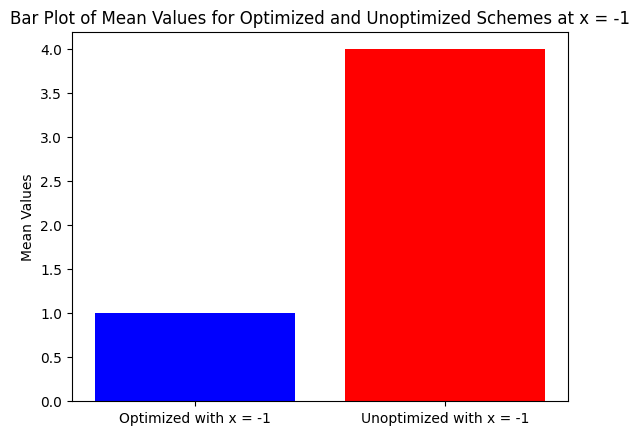}
    \includegraphics[width = 0.675\linewidth]{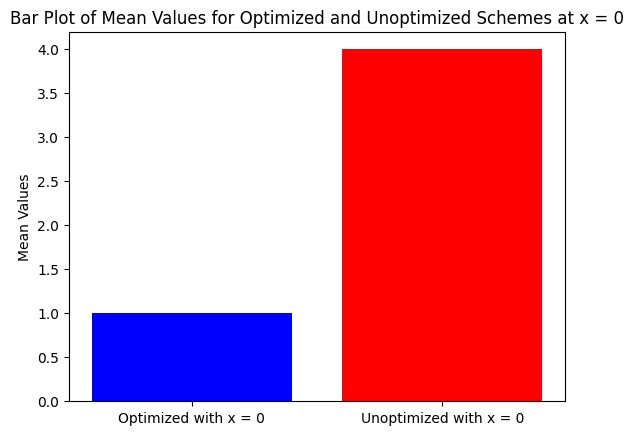}
    \caption{Average of REBA scores between Unoptimized and Optimized Schemes Depending on Starting Positions}
    \label{rebresutls}
\end{figure}
As it is shown in Figure \ref{rebresutls}, we can see a universal decrease in REBA scores in both starting locations using the ergonomically optimized scheme as opposed to a naive approach.
\section{Discussion}

One of the most prominent obstacles to practical ergonomic optimization for intelligent applications such as pHRI can be attributed to the existing metrics with mathematical complexities. REBA calculates the ergonomic risk mainly through ranges of the angle of the specific joint in a step-wise manner. Most ergonomic metrics have two (usually contradictory) qualities: generality and accuracy \cite{fransson1995portable}. The former refers to the metric's capacity to be applicable across various tasks, contexts, and rates of operation, signifying its universality. The latter pertains to the metric's exactness in quantifying the genuine ergonomic hazards associated with a particular task as determined by the framework. Recent sensitivity analyses indicate that the REBA methodology exhibits limited suitability for material handling tasks \cite{REBAbad}, which constitute a significant aspect of pHRI object handovers. Instead, its efficacy is more pronounced in interventions targeting overall postural ergonomics. The speed and ease of use have been the strongest catalyst to REBA's popularity in the industry since its inception in the year 2000. However, the technological constraints of the past obscured the inadequacies of REBA for mathematical optimization tasks. Now, with advancements in image processing and sensor accuracy, we can gather and interpret gait data — an application perhaps not originally intended by REBA's creators.
\par
We encounter these shortcomings in our experiments. We use the tables given by Figure \ref{fig:worksheet} to calculate the REBA score of the participants. The REBA method has two separate scores that need to be calculated which then can be used through a series of tables which results in a final REBA score. In the worksheet, Score A is related to the Neck, Trunk, and Leg Analysis, and Score B is related to the "Arm and Wrist Analysis'. For the majority of the cases, at initial position x=0, we calculate a score of 2 for Score A and a score of 5 for Score B. By using Table C in the worksheet and accounting for the activity score, we end up with a REBA score of 4. For initial position x= -1, the framework returns a score of 3 for Score A and a Score of 5 for Score B. This main difference between the REBA scores of the two initial positions is due to the twisted trunk induced by the asymmetry of the final object placement concerning the body by the scheme. However given Table C, we still result in a final REBA score of 4. However, there is a significant difference in comfort and physical strain between the two iterations. This difference is somewhat represented in the simple summation of the postural score but not inside the final valuation of the REBA score. This further demonstrates the shortcomings of REBA and the benefit of using more granule scoring schemes such as the postural scores presented in this paper which are derived from REBA itself.
\par
We can see as shown in the results section, that the postural score serves as an adequate optimization goal for this method. This was to be expected since the postural score is REBA in a more granule form. VR and computer simulations have the luxury of being able to calculate high-precision data with ease. The postural score approach also grants more flexibility in scoring each limb differently. By not limiting itself to the table presented in REBA, developers can create personalized scores and optimizations depending on the specific needs of the worker. Scores can be assigned with different numbers and values depending on the nature of the task and the health history of the individual. However, this approach demands further investigation to ensure its scientific validity and real-life efficacy. This further demonstrates the need for mathematical approaches to improve ergonomic optimization outside the scope of REBA (as demonstrated in this study) for not only pHRI contexts but also potential applications in VR. Additionally, more specialized and sensitive frameworks for pHRI applications in industrial settings could be devised. 
\par
\section{Conclusion}
The implications of the findings are paramount for the pHRI practice in industries such as construction, manufacturing, and retail warehousing with substantial fieldwork. Robots are introduced to industrial applications to increase the safety of manual work, among others, through automating and taking over labor-intensive and repetitive tasks that are often not ergonomic. While their presence in the field should not pose additional safety hazards (e.g., struck-by accidents), their operations should also guarantee worker safety. As the only comprehensive framework to assess ergonomics of body posture, REBA has features that limit its functionality for real-time, feedback-enabled mechanisms needed for on-spot robot perception and reaction. As such, the developed methodology enhances the applicability of REBA for adaptive, safe, and fast implementation of REBA in bi-manual material handover situations between robots and field workers.
\par
Another benefit of this approach stems from the fact that the robot can learn personalized traits outside of operating hours. Once the robot has gained the relative position for a specific task with the specific worker, it can perform ergonomically optimized pHRI object handover with or without the Unity game engine. The learner optimizes using pixel units inside the game engine. These ratios are simply a change of dimensions relative to the body measurements of the worker imported into the 3D environment, resulting in a representation of body area and volume in a digitized setting. Subsequently, the real-world implementation would just consist of a dimension transformation for the robot to perform the movement with or without the game engine. Current robotic research has enabled developers to have an accurate visual representation of localization and dimension perception through LiDAR technology \cite{queralta2019fpga}, stereo vision \cite{montoya2022assisted} and SLAM-based algorithms \cite{TAHERI2021104032} for a variety of different applications. Due to this measurement strategy, it is of paramount importance that the worker 3D humanoid model and its associated IK framework be as representative as possible of the worker's true body mechanics to decrease the potential divergence of the training and interaction optimality.
\par

From a robotic implementation standpoint, this framework would require the exact knowledge of the global positions of the worker and the robot. Since the RL scheme returns an optimized position relative to the worker, it is necessary to know where the person is to transport the box to the relative position. Previous works have shown that worker localization via Indoor GPS can be achieved with high accuracy \cite{KHOURY2009444}. Such methods can be implemented in potential frameworks in order to enable the RL scheme in an applied setting. 
\par
The VR training method also opens up the groundwork for expedited framework testing outside of the job site. By testing pHRI frameworks and interactions in a virtual environment, we reduce the risk of potentially hazardous failure in a setting where humans are present. Many edge environmental edge cases can be analyzed and simulated which would pose a potential risk in real deployments.  

\par
While this research presents a novel approach toward using ergonomics safety measures in pHRI, there are a few limitations that can be addressed in future work. The reward policy can be modified to better guide the Q-learner to reach the optimal position. Rewarding different features such as body ratios and joint angles can be able to assist the Q-learner's convergence. However, this would discourage the current model-free nature of the learner, since we are guiding it to the ideal position of the task by adding task-specific information into the training regiment. The boundary space can also be further limited to encourage faster convergence. Exploring the entire boundary space took a significantly longer duration compared to the RL algorithm. By further truncating the possible solution spaces, the odds of converging to a local minima decrease, and the higher probability of the RL algorithm reaching the global optima faster, especially given the fact the softmax action selector's temperature has been set to a high value. However limiting the boundary would decrease the routine's generalizability due to the reason that given different tasks, configurations, and worker demographics, the REBA distribution can be highly variable. More sophisticated RL algorithms can be implemented for faster convergence. For future work, we plan to incorporate Deep Q-learning methods \cite{fan2020theoretical} which utilize neural networks to represent the Q-function as opposed to the table of Q-values, and methods such as Soft Actor-Critic algorithms \cite{haarnoja2018soft}. However, the biggest challenge to overcome is, once again, the discrete nature of the ergonomic frameworks.
\par
Although the IK framework does provide an adequate representation of postural kinematics, a gap between true kinematics and the simulated ones is to be expected. We selected Unity for simulating kinematics over other scientific simulators because we aim to utilize this game engine in future HRI projects. Specifically, we intend to use Unity for robot logic and to integrate VR and AR for remote work and worker training applications. The objective of our approach is not to replicate human postural kinematics with high fidelity but rather to establish and refine a framework that is informed by ergonomic principles for use in HRI contexts. Options such as OpenSim would enable to model and simulate the bio-mechanics and ergonomics of humans to a much more precise scale. However, scientific simulators and their integration with robotic controls and generalizability in a virtual environment or a construction job site can prove challenging.
\par
\begin{figure}[H]
    \centering
    \includegraphics[width = 0.45\linewidth]{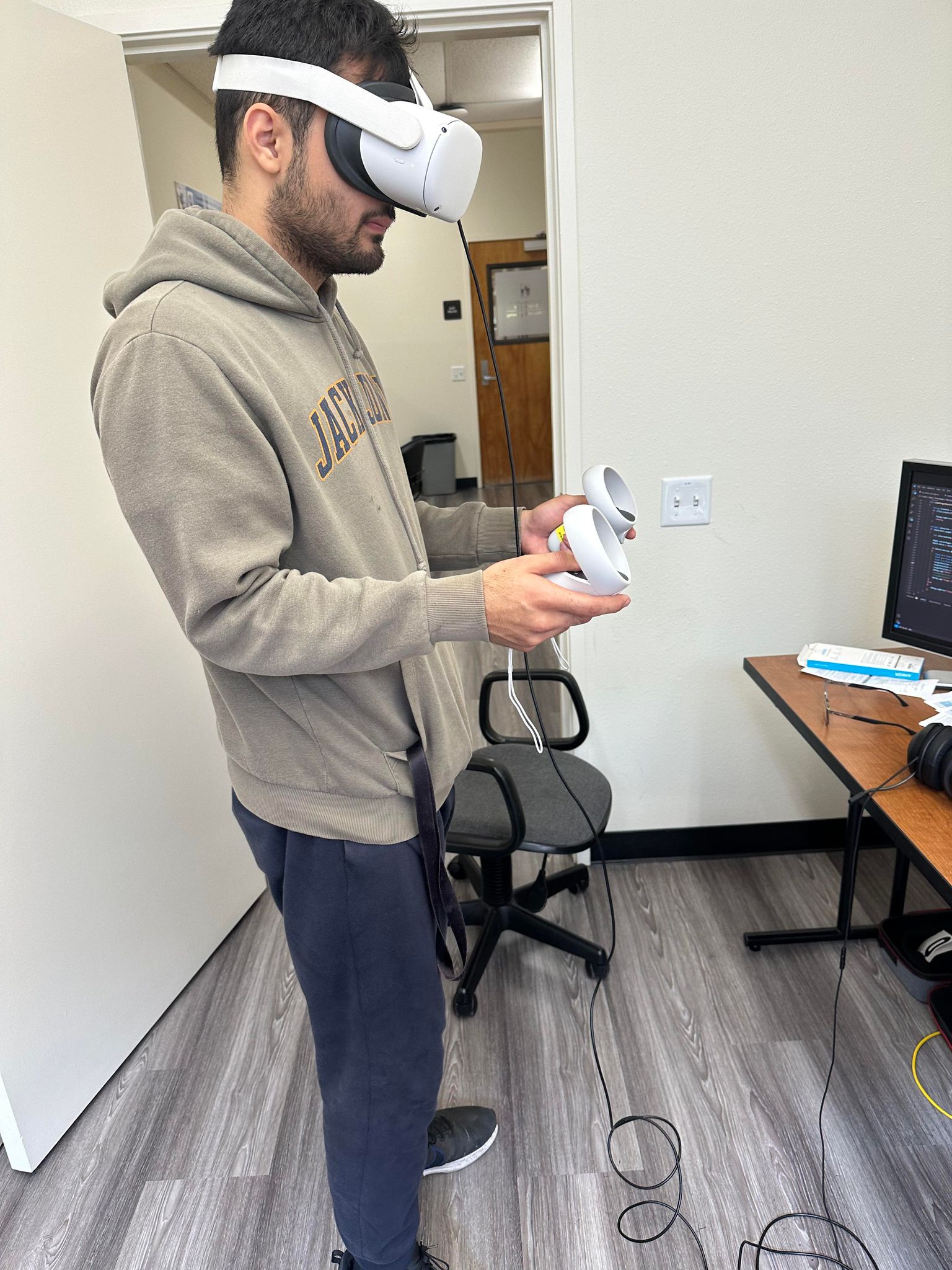}
    \includegraphics[width = 0.45\linewidth]{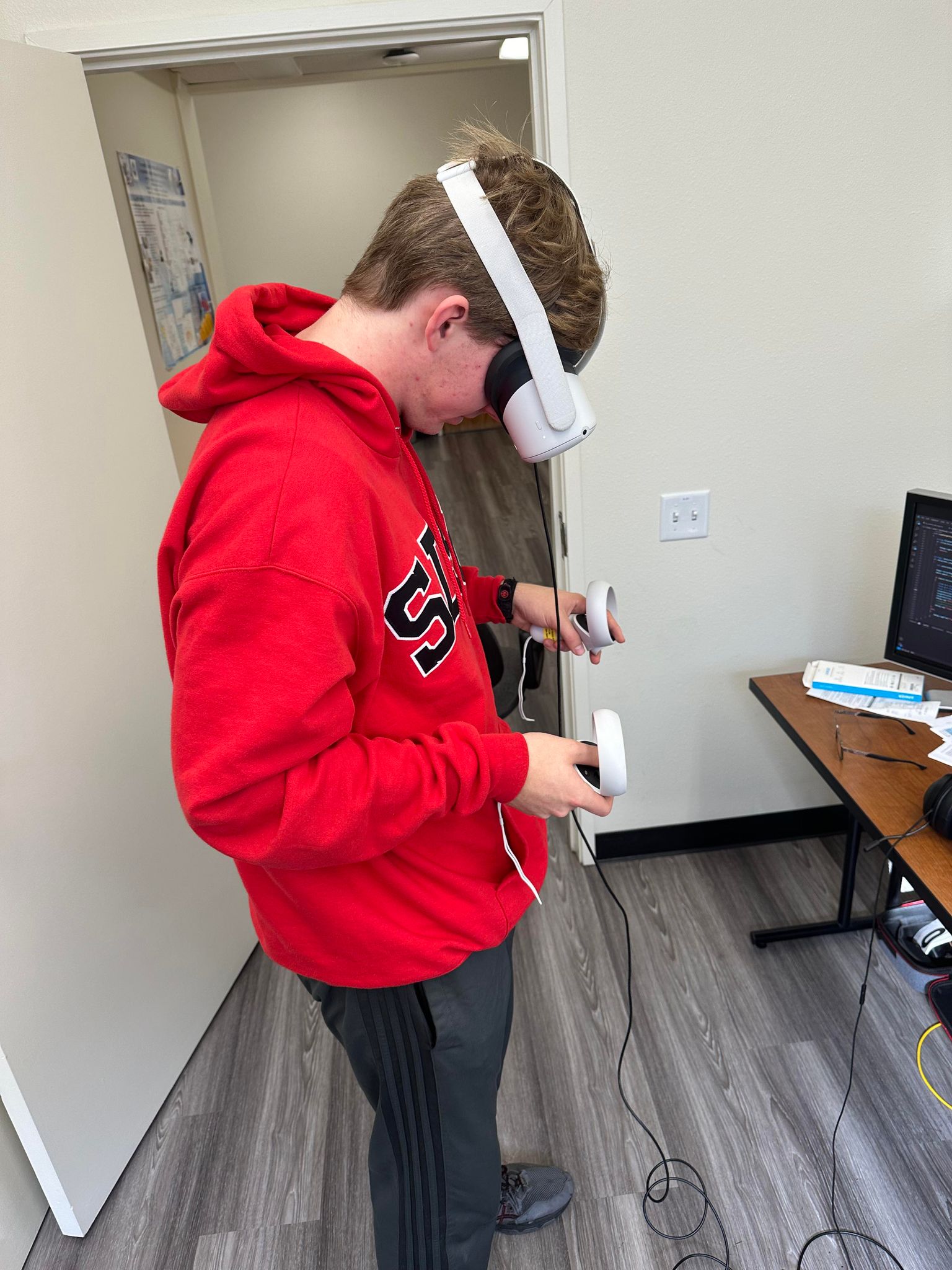}
    \caption{Difference in Manifested Ergonomic Posture on Human Subjects when Using the Identical RL Optimized Location}
    \label{diff}
\end{figure}

For the experiment, the humanoid model in the game engine was not tailored to every single participant. We expect that this factor has reduced the maximum benefits that the regiment has to offer. However, this decision was deliberate since we were planning to investigate the generalizability of the scheme. We still were able to improve ergonomic scores while also achieving identical REBA scores among all participants These results show a significant level of generalizability. However, we predict that these scores are not the true optimal positions for every participant. It would be ideal to personalize the humanoid model to every construction worker in an applied setting for maximum ergonomic optimization. The discrepancy in real-life manifestations of the human posture can be seen in Figure \ref{diff}. Even though there is a difference between the qualitative postures, this method still provides significant benefits as compared to ergonomically agnostic heuristics. Furthermore, such positions outside of the scope of the study can have either exact or marginally different REBA scores. However, it is not hard to infer that these postures can have different comfort levels and injury risks. This is another piece of evidence that points towards the need for sensitive and mathematical ergonomic and safety frameworks for pHRI tasks.
\par

Finally, this object handover method is considered a non-adaptive approach \cite{kappler2023optimizing}. A non-adaptive approach means that the robot simply moves the object to a pre-defined path without any adaptation to the potential changes of the receiver. For future work, we plan to incorporate adaptive corrections to the object handover for stronger generalizability in a dynamic environment which is paramount to scalable deployment of construction robots.
For future work, we would like to test this framework in a dynamic construction environment using physical robots. By defining the task and the human model of the construction worker. VR has been previously used to assist construction workers in training for HRC \cite{VRConstructionGerber}. This work further opens up the possibilities of utilizing VR to tackle the pressing construction problems. One major obstacle to the deployment of construction robots can be attributed to the lack of trust in the feasibility and efficacy of the robots. Pairing training and ergonomic considerations inside and outside of the job site. 

\section*{Acknowledgements}
The presented work has been supported by the U.S. National Science Foundation (NSF) CAREER Award through grant No. CMMI 2047138, as well as a scholarship funded by grant No. DUE 1930546. The authors gratefully acknowledge the support from the NSF. Any opinions, findings, conclusions, and recommendations expressed in this paper are those of the authors and do not necessarily represent those of the NSF. 
The authors would also like to thank Dr. Nguyen-Truc-Dao Nguyen, Dr. Miguel Dumett, and Mr. Jeffrey Xing for their advice and support as well as our colleagues at the Data-informed Construction (DiCE) Lab at San Diego State University for their voluntary assistance in the evaluation of the framework. 
\section* {Conflict of Interest}
The authors identify no conflict of interest.
\section*{Data Availability}
Some or all of the code and data are available upon request. 

\section*{CreDiT authorship contribution}
\textbf{Mani Amani}: Conceptualization, Data Curation, Methodology, Investigation, Software, Validation, Visualization, Writing - Original Draft, Writing- Review and Editing
\par
\textbf{Reza Akhavian}: Funding Acquisition, Project Administration, Conceptualization, Supervision, Writing- Review and Editing

 \bibliographystyle{elsarticle-num} 
 \bibliography{cas-refs}





% \end{thebibliography}
\end{document}